\documentclass{article}
\usepackage[latin9]{inputenc}
\usepackage{wrapfig}
\usepackage{amsmath}
\usepackage{amssymb}
\usepackage{graphicx}
\usepackage[authoryear]{natbib}
\usepackage[unicode=true,
 bookmarks=false,
 breaklinks=false,pdfborder={0 0 1},backref=section,colorlinks=false]
 {hyperref}

\makeatletter
\newenvironment{lyxcode}
	{\par\begin{list}{}{
		\setlength{\rightmargin}{\leftmargin}
		\setlength{\listparindent}{0pt}
		\raggedright
		\setlength{\itemsep}{0pt}
		\setlength{\parsep}{0pt}
		\normalfont\ttfamily}%
	 \item[]}
	{\end{list}}

\usepackage[preprint]{neurips_2019}

\@ifundefined{showcaptionsetup}{}{%
 \PassOptionsToPackage{caption=false}{subfig}}
\usepackage{subfig}
\makeatother

\begin{document}

\title{Controlling Covariate Shift using\\
Balanced Normalization of Weights }

\author{Aaron Defazio \& L\'eon Bottou\\
Facebook AI Research New York}
\maketitle
\begin{abstract}
We introduce a new normalization technique that exhibits the fast
convergence properties of batch normalization using a transformation
of layer weights instead of layer outputs. The proposed technique
keeps the contribution of positive and negative weights to the layer
output balanced. We validate our method on a set of standard benchmarks
including CIFAR-10/100, SVHN and ILSVRC 2012 ImageNet.
\end{abstract}

\section{Introduction}

The introduction of normalizing layers to neural networks has in no
small part contributed to the deep learning revolution in machine
learning. The most successful of these techniques in the image classification
domain is the \emph{batch normalization} (BatchNorm) layer \citep{batchnorm},
which works by normalizing the univariate first and second order statistics
between layers. 

Batchnorm has seen near universal adoption in image classification
tasks due to its surprisingly multifaceted benefits. Compared to an
unnormalized network, its has been widely observed that using batch
norm empirically results in:
\begin{itemize}
\item Stability over a wide range of step sizes
\item Faster convergence (particularly with larger step sizes)
\item Improved generalization
\end{itemize}
The multiple effects of BatchNorm make it both hard to replace and
hard to analyze. In this paper we introduce Balanced Normalization
(BalNorm), a normalization that works in weight space and still uses
a form of batch statistics unlike previous weight space approaches.
BalNorm results in very rapid convergence, even more so than BatchNorm,
however as we will show in our experiments, this also results in a
tendency to overfit. When combined with additional regularisation,
BalNorm can significantly outperform BatchNorm, which benefits less
from this additional regularisation. 

\section{Related Work}

A number of normalization layers have been proposed that can be considered
alternatives to batch normalization. Batch normalization has also
been extended as batch renormalization \citep{batchrenorm} to handle
smaller batch sizes.

\paragraph*{Layer/Instance Normalization}

A simple modification of BatchNorm involves computing the statistics
independently for each instance, so that no averaging is done across
each mini-batch, instead averaging either across channels (layer norm)
or separately for each channel (instance norm). Unfortunately these
techniques are known to not generalize as well as batch norm for convolutional
neural networks \citep[Sec 6.7; Sec 4.1.][]{layernorm,groupnorm}.

\paragraph*{Group Normalization}

A middle ground between layer and instance normalization can be found
by averaging statistics over small groups of channels. This has been
shown empirically to be superior to either approach, although there
is still a gap in generalization performance \citep{groupnorm}. Like
the approaches above, it avoids a dependence on batch statistics,
allowing for potentially much smaller batches to be used without a
degradation in generalization performance. 

\paragraph*{Weight Normalization}

Additional stability can be introduced into NN training by constraining
the norm of the weights corresponding to each output channel/neuron
to be one. When this is done by an explicit division operation in
the forward pass, rather than via an optimization constraint, this
is known as weight normalization \citep{weightnorm}. An additional
learnable scaling factor is also introduced. Unfortunately, to match
the generalization performance of BatchNorm on image classification
tasks such as CIFAR-10, this technique needs to be used together with
partial (additive only) BatchNorm \citep[Section 5.1,][]{weightnorm}.

\paragraph*{Local Response Normalization}

A precursor to batch norm, local normalization methods \citep{lecunobject,divisivenorm}
played an important part in the seminal AlexNet architecture \citep{alexnet},
and were widely used before batch norm was introduced. LR normalization
has similarities to group norm in that it uses a group of neighboring
channels (with ordering set arbitrary at initialization) for normalization.
Although it aids generalization in a similar manner to BatchNorm,
it does not accelerate convergence or allow for larger step sizes
to be used \citep[Sec 4.2.1,][]{batchnorm}.

\section{Assumptions}

The BalNorm method functions by modifying the weights of a convolution
before it is applied. For justifying the form of our method, we make
the following assumptions about this convolution, which we will discuss
relaxing after detailing the method:
\begin{enumerate}
\item All inputs to the convolutional layer are positive, such as when the
layer is preceded by a ReLU.
\item The convolution has stride one.
\item Cyclic padding is used.
\item All weights are non-zero, and there exists at least one positive and
one negative weight per output channel.
\end{enumerate}

\section{Method}

Consider initially for simplicity a convolutional layer with a single
input and output channel. Let 
\[
w:\mathtt{kernelheight}\times\mathtt{kernelwidth},
\]
 be the weight kernel for this neuron, and let 
\[
x:\mathtt{batchsize}\times\mathtt{height}\times\mathtt{width},
\]
be the input tensor for a single mini-batch. We will compute scalar
quantities $s$ and $b$ that modify the weights as follows:
\[
w^{\prime\prime}=sw^{\prime}=s\left(w+b\right).
\]
This transformation will be fully differentiated through during the
backwards pass (using automatic differentiation) so that the gradient
of $w$ is correct. As with BatchNorm, we also include an additional
affine transformation after the convolution to ensure no expressivity
is lost due to the normalization operation.

The core idea of balanced normalization is to balance the contribution
of positive and negative weights to the output of the convolution.
To this end, we introduce additional notation to address the positive
and negative weights separately. Let superscripts $+/-$ (i.e. $w^{+}/w^{-})$
indicate sums of the positive/negative elements respectively. Also,
let $v$ be the sum of the input data to the layer,
\[
v=\sum_{i,j,k}x_{ijk}.
\]
As we have two constants to determine, we need two constraints that
we wish to be satisfied. The first constraint we introduce is common
with batch normalization, a constraint on the mean of the output.
Since under the cyclic padding assumption, each weight is multiplied
by each input element, we can constrain the mean of the output to
be zero by requiring that:
\[
vs\sum_{j,j}\left(w_{jk}+b\right)=0,
\]
\[
\therefore b=-\text{mean}(w).
\]
The second constraint controls the magnitude within the total output
of the layer, of the positive weight elements:
\[
vsw^{\prime+}=r,
\]
\[
\therefore s=\frac{r}{vw_{+}^{\prime}}.
\]
The constant $r$ is set so that the contribution is of average $1$
per output element, which is achieved by setting $r$ to the product
of batch-size, output width and output height. Note that due to the
mean constraint, this automatically results in the negative weight
contribution also being of magnitude $r$.

\subsection{Full case}

When multiple input channels are used, each weight no longer multiples
each input, rather they each multiply only inputs from a single channel.
To compensate for this we need to compute per-channel sums $v_{c}$
(where $c$ is the channel index) and change the second constraint
as follows:
\[
\sum_{c}^{\mathtt{channelsin}}v_{c}sw_{c}^{\prime+}=r.
\]
The first constraint changes in the same fashion. 

When multiple output channels are used, we just duplicate this procedure
applying it to each channel's weights separately. We thus maintain
a $s$ and $b$ value per output channel, and compute as intermediate
values a $w^{\prime+}$ of matrix shape. For completeness we give
the full equations below. All summations are over the full range of
the summed indexes.

\paragraph*{Tensor shapes
\begin{align*}
w: & \mathtt{channelsout}\times\mathtt{channelsin}\times\mathtt{kernelheight}\times\mathtt{kernelwidth}\protect\\
x: & \mathtt{batchsize}\times\mathtt{channelsin}\times\mathtt{heightin}\times\mathtt{widthin}\protect\\
w^{\prime+}: & \mathtt{channelsout}\times\mathtt{channelsin}\protect\\
v: & \mathtt{channelsin}\qquad s,b:\mathtt{channelsout}
\end{align*}
}
\begin{lyxcode}
Updates:
\begin{eqnarray*}
v_{c} & = & \sum_{i,j,k}x_{icjk},\qquad w_{dc}=\sum_{j,k}w_{dcjk},\\
r & = & \mathtt{batchsize}\times\mathtt{heightout}\times\mathtt{widthout}\times\mathtt{stride}^{2}\\
b_{d} & = & -\frac{\sum_{c}v_{c}w_{dc}}{\left(\mathtt{kernelheight}\times\mathtt{kernelwidth}\right)\sum_{c}v_{c}},\\
w_{dc}^{\prime+} & = & \sum_{j,k}\left(w_{dcjk}+b_{d}\right)I[w_{dcjk}+b_{d}>0],\\
s_{d} & = & \frac{r}{\sum_{c}v_{c}w_{dc}^{\prime+}},\\
w_{dcjk}^{\prime\prime} & = & s_{d}\left(w_{dcjk}+b_{d}\right).
\end{eqnarray*}
\end{lyxcode}
At test time, we follow the technique used in BatchNorm of using a
running estimate of the data statistics ($v_{c}$ in our method) that
is computed during training time. 

\subsection*{Single pass formulation}

The above calculation requires two passes over the weights, first
to compute $b$, then to compute the sum of positive elements after
the addition of $b$. We can do an approximate computation using only
one pass by assuming the sign of each element does not change after
the addition of $b$. The $s$ calculation changes as follows:
\begin{eqnarray*}
n_{dc}^{+} & = & s_{d}\sum_{j,k}I[w_{dcjk}>0],\quad s_{d}=\frac{r}{\sum_{c}v_{c}\left(w_{dc}^{+}+b_{d}n_{dc}^{+}\right)}.
\end{eqnarray*}
We use this variant in all experiments that follow. 
\begin{figure}
\centering{}\includegraphics[scale=0.65]{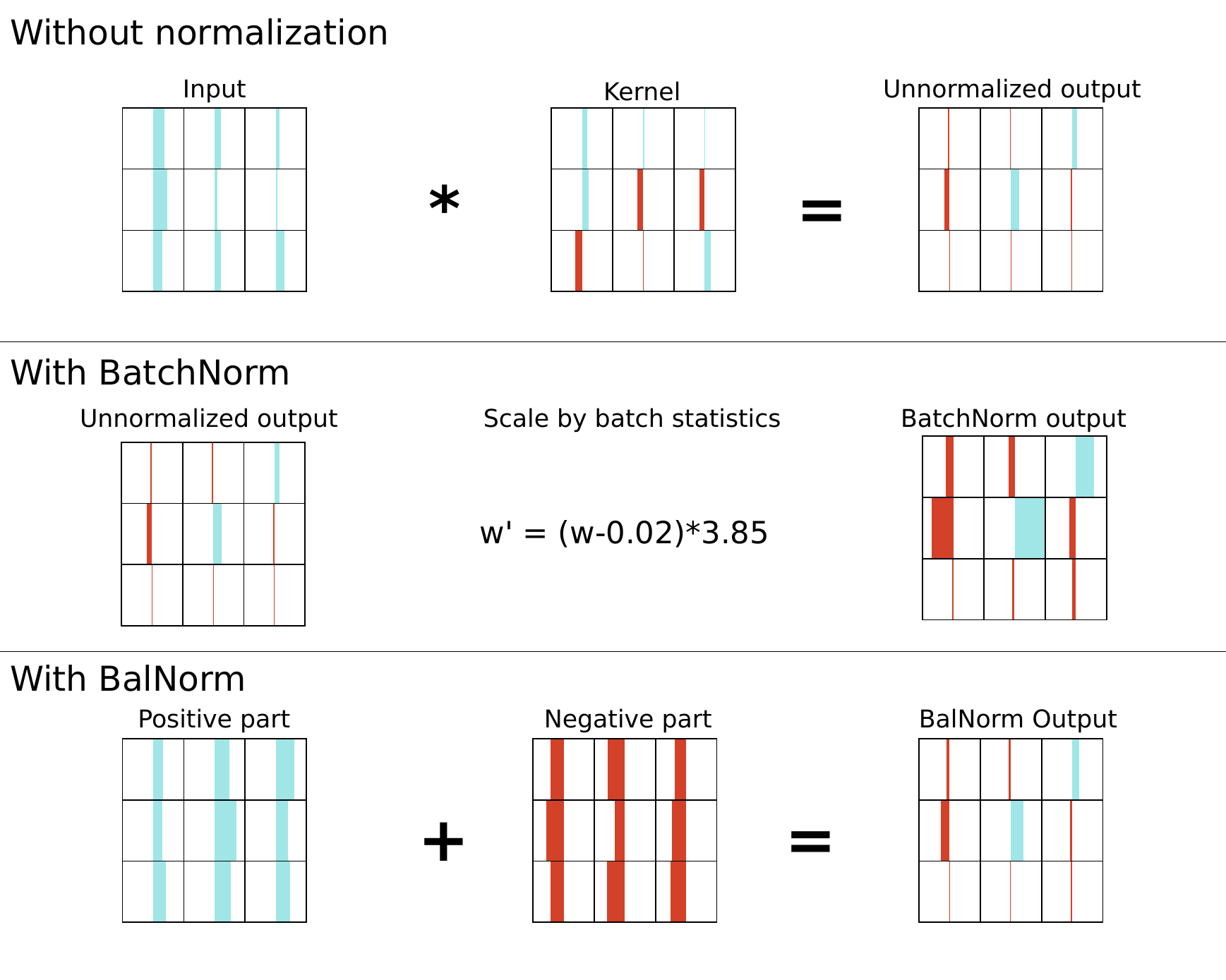}\caption{Balanced normalization ensures the contribution from positive and
negative kernel weights to the output remains of the same total magnitude,
both compared to each other and between epochs. In the case shown
of a $3\times3$ kernel (cyclic convolution) against a $3\times3$
image with padding $1$, this magnitude is $\text{width}_{\text{out}}\cdot\text{height}_{\text{out}}=9.0$.}
\end{figure}

\section{Discussion}

The original justification for batch normalization is its ability
to minimize covariate shift, although there is some debate on whether
or not this is the main contributing factor to its effectiveness,
see \citet{mitbnstats}. In this context, covariate shift refers to
the change between steps of the statistics of the outputs of a layer.

Like batch normalization, the approach we propose also controls the
shift in the outputs of a layer between steps, just in a different
way. Our approach is motivated by a hypothesis that it is not necessary
to control the mean and variance precisely; other notions of scale
and shift may work as well or better. \citet{mitbnstats} show that
normalizing by other norms, such as $L_{1}$, can work well, supporting
this hypothesis.

The sum of the output from positive and negative weights is a form
of $L_{1}$ control which can be contrasted with the $L_{2}$ control
that Batchnorm uses. This control can be motivated by Young's convolution
inequality, which bounds the output of a convolution operation in
terms of the norm of the input:
\[
\left\Vert x*w\right\Vert _{r}\leq\left\Vert w\right\Vert _{p}\left\Vert x\right\Vert _{q},
\]
\[
\text{where }\frac{1}{p}+\frac{1}{q}=\frac{1}{r}+1.
\]
Note that $p,q,r\geq1$ is also required, and that this only applies
directly when there is a single input and output channel, which we
assume in the remainder of this section for simplicity. 

For BalNorm, we have assumed that the input is positive, so that our
input sum is equivalent to the $L_{1}$ norm of the input. Additionally,
after subtracting off the mean, the weight vector $w^{\prime}$ has
$L_{1}$ norm equal to $w^{\prime+}-w^{\prime-}=2w^{\prime+}$, so
we are also normalizing the weights by the $L_{1}$ norm. In effect,
we are applying Young's convolution inequality with $p=q=r=1$. 

It is also possible to apply the above inequality with $p=2$, $q=1$
and $r=2$. I.e. normalize the weights using the $L_{2}$ norm, giving
a bound on the $L_{2}$ norm of the output in terms of the $L_{1}$
norm of the input. This is less satisfying as one convolution's output
is the input of another convolution (after passing through scaling
\& a nonlinearity) so we would like to use the same norm for both
inputs and outputs. The related weight normalization \citep[WN,][]{weightnorm}
method normalizes weights by their $L_{2}$ norm, and differs from
our method by centering outputs using an additional mean-only output
batchnorm. Additionally, since it doesn't normalize by the input norm,
the output norm can be correspondingly large. These differences have
a significant effect in practice. 

\subsection{Assumptions}

\subsubsection*{Input to the convolutional layer is positive}

This assumption is not necessary for the implementation of our method,
rather it ensures that the output is more constrained than it otherwise
would be. When ReLU nonlinearities are used in the standard fashion,
this assumption holds except in the first layer of the network where
input pixels are usually in the range {[}-1,1{]}, due to pre-normalization.
We recommend this pre-normalization is removed, as it is unnecessary
when normalization happens immediately inside the first convolution.
Recommendations in the literature that suggest input normalization
is beneficial are usually referring to networks without per-layer
normalization.

\subsubsection*{Non-strided convolutions}

A strided convolution can thought of as a non-strided convolution
with the extra output values thrown away. If Balanced normalization
is used with a strided convolution, the contribution to the output
from positive and negative weights will no longer be exactly balanced.
In practice the violation will be small if the output of the non-strided
version of the convolution is smooth. 

\subsubsection*{Cyclic padding }

Our equations for $b$ and $s$ assume that each weight for an input
channel is multiplied by each input value for that channel. Most deep
learning frameworks use zero-padded convolutions instead of cyclic
padding, which violates this assumption. In practice we do not find
this violation to be troublesome, as it only affects edge pixels,
and has a dampening effect as it only reduces the output contribution
of the positive or negative weights.

\subsubsection*{All weights are non-zero, and there exists at least one positive
and one negative weight per output channel}

We avoid the use of an $\epsilon$ parameter such as used in BatchNorm,
as the denominator of our normalization factor is only zero if \emph{every}
weight for every input channel is simultaneously positive (or all
negative), or the weights become extremely small. The later case does
not appear to happen in practice. Nevertheless, we find it helps to
initialize the weights in a balanced fashion, so that no channel's
weight kernel is all positive or all negative. We do this by modifying
the default initialization by resampling any such kernel-weight's
signs.

\section{Experiments\label{sec:experiments}}

In our plots we show a comparison to BatchNorm and GroupNorm. We omit
a comparison to Layer/Instance Normalization as our initial experiments
were consistant with findings in the literature that show that they
are inferior to GroupNorm and BatchNorm, at least for the convolutional
architectures we consider below \citep{groupnorm}. We also performed
a comparison against the WeightNorm method, however despite significant
efforts we were not able to get it to reliably converge when using
very deep architectures such as ResNet-152 that we choose for our
experiments. We could not find any results in the literature where
it is sucessfully applied to state-of-the-art deep networks, and we
believe this is a real limitation of the method.

\subsection{CIFAR-10/100}

The CIFAR-10 dataset \citep{cifar} is considered a standard benchmark
among image classification tasks due to its non-trivial complexity,
requiring tens of millions of weights to achieve state-of-the-art
performance, and also its tractable training times due to its small
size (60,000 instances). The downside of this small size is that significant
care must be taken to avoid overfitting. This overfitting can be partially
avoided using data augmentation, and we followed standard practice
of using random horizontal flips and crops (pad 4px and crop to 32px)
at training time only. \begin{wrapfigure}{r}{0.45\columnwidth}%
\centering{}\includegraphics[scale=0.9]{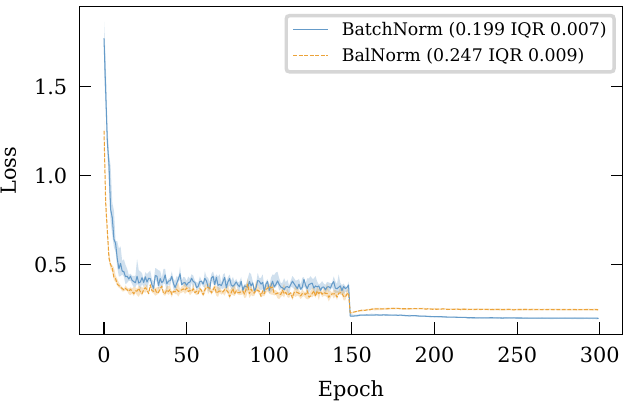}{\small{}\caption{{\small{}\label{fig:overfitting}Indications of overfitting, as test
loss starts to increase significantly after the first learning rate
decrease.}}
}\end{wrapfigure}%

Our initial experiments involving a non-bottleneck wide ResNet network
with 20 convolutions, 96 initial planes after first convolution and
9.7m parameters. The hyper-parameters used were LR: 0.1, SGD+Mom:
0.9, decay: 0.0001, batch-size: 128, 1 GPU, 10 fold learning reductions
at epochs 150 and 225, and standard fan\_out normal initialization
following \citet{delvingdeep}. These parameters are defaults commonly
used with BatchNorm and were not tuned.

Our experiments indicated that our BalNorm approach converged significantly
faster than BatchNorm, but also overfit significantly more ({\small{}Figure
\ref{fig:overfitting}).} 

We believe this is caused by the batch statistics having less noise
with BalNorm than BatchNorm, rather than the faster initial convergence,
as experiments involving reduced step sizes did not further improve
generalization. Similarly, we were not able to achieve comparable
fast initial convergence by using larger step-sizes with BatchNorm. 

We found instead that we could match the generalization of BatchNorm
using either of the following two approaches:
\begin{enumerate}
\item Using fewer instances to compute the batch statistics. Using the first
quarter of the batch to compute the statistics used for the full batch
successfully fixed the overfitting seen in Figure \ref{fig:overfitting},
resulting in higher test accuracy for BalNorm (95.8\%) over BatchNorm
(94.7\%) and no overfitting visible in test loss.
\item Using mixup \citep{mixup} or manifold mixup \citep{manifold-mixup},
which introduce activation noise of a similar nature. 
\end{enumerate}
We recommend the use of manifold mixup, as it significantly improves
test accuracy for both BatchNorm and BalNorm, although it does result
in higher test loss in some cases.

Following closely the approach of \citealt{manifold-mixup}, we applied
both BalNorm and BatchNorm to the larger near state-of-the-art pre-activation
ResNet-152 architecture (\citealt{preact}, 58.1m parameters, {[}3,8,36,3{]}
bottleneck blocks per layer respectively, 64 initial channels), using
a modified version of their published code and the hyper-parameters
listed above, with manifold mixup used for each method. As Figure
\ref{fig:CIFAR-10} shows, the test set performance is essentially
the same at the final epoch, but BalNorm converges significantly faster
at the early epochs.

\subsubsection*{CIFAR100}

We also achieved a similar performance on the CIFAR-100 dataset (which
has similar properties to CIFAR10) as shown in Figure \ref{fig:CIFAR-100},
where we used the same hyper-parameters and network architecture as
for CIFAR10.

\subsection{Shorter duration training}

\begin{wrapfigure}{r}{0.45\columnwidth}%
\centering{}\includegraphics[scale=0.9]{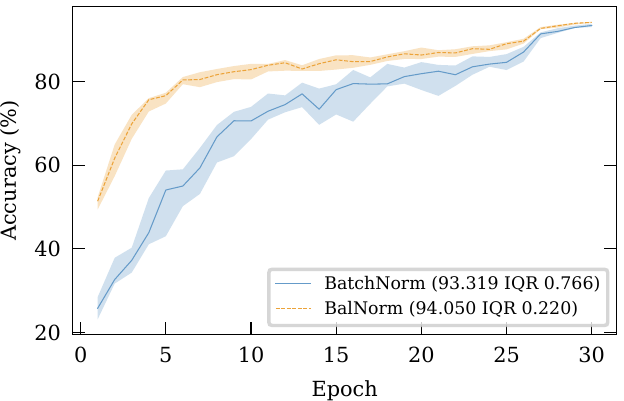}{\small{}\caption{{\small{}\label{fig:30_epochs}Shorter duration CIFAR10 training (WRN
network)}}
}\end{wrapfigure}%

Given the encouraging results above during the early stages of optimization,
we investigated if BalNorm was superior when training is restricted
to 30 epochs instead of 300. We used a ``super convergence'' learning
rate schedule as suggested by \citet{superconvergence}, consisting
of a 5 fold ramp in learning rate (starting at 0.1) from epochs 1
to 13, then a 5 fold ramp down to epoch 26, followed by further annealing
by 100x down over the remaining epochs. Momentum follows a reverse
pattern, from 0.95 to 0.85 to 0.95, and fixed at 0.85 after epoch
26. Manifold mixup was used again for both methods. Using this schedule
BalNorm shows a 94.0\% (IQR 0.22) median test accuracy compared to
93.3\% for BatchNorm (IQR 0.77).

\subsection{Street View House Numbers}

\begin{figure}
\begin{centering}
\subfloat[\label{fig:CIFAR-10}CIFAR-10 (PreResNet152 with manifold mixup $\alpha=6$)]{\includegraphics{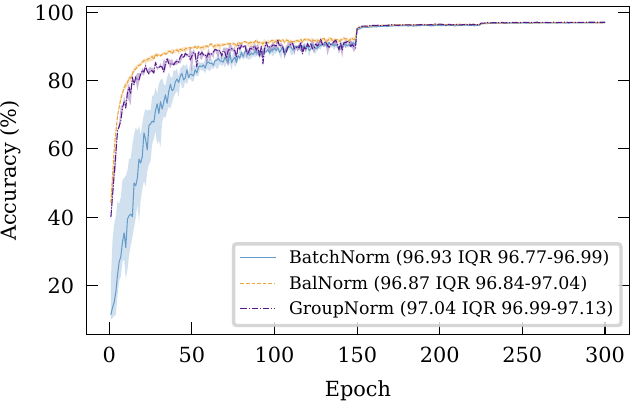}

\includegraphics{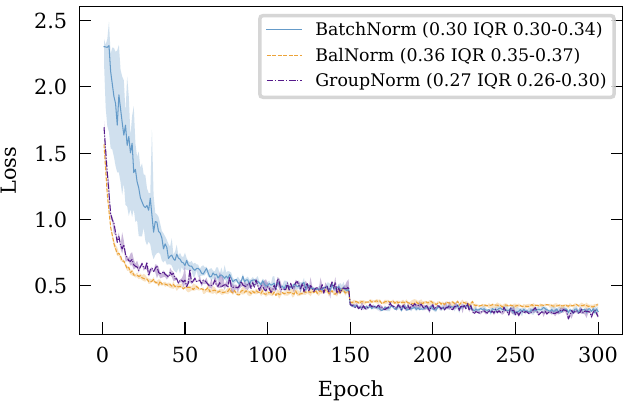}}
\par\end{centering}
\begin{centering}
\subfloat[\label{fig:CIFAR-100}CIFAR-100 (PreResNet152 with manifold mixup
$\alpha=6$)]{\includegraphics{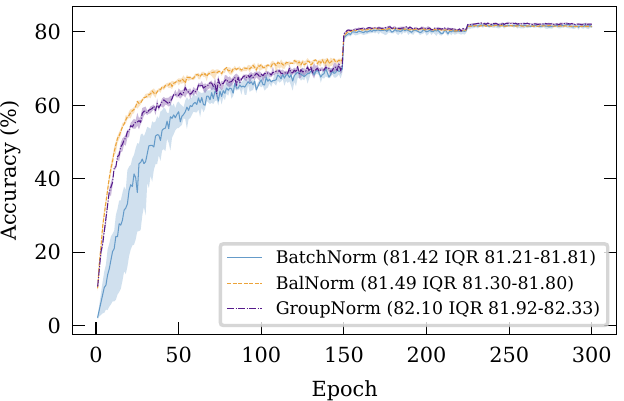}

\includegraphics{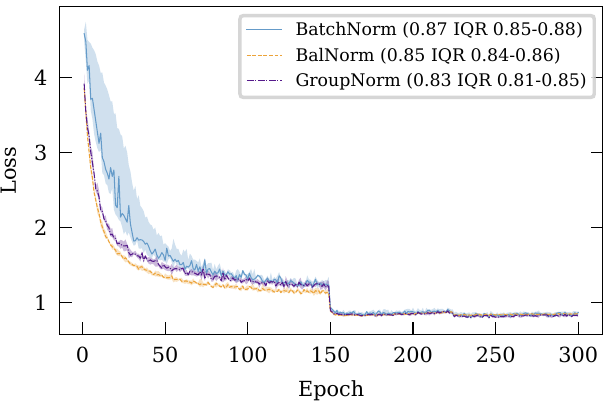}}
\par\end{centering}
\begin{centering}
\subfloat[\label{fig:SVHN}Street View House Numbers + extra (WRN with manifold
mixup $\alpha=6$)]{\includegraphics{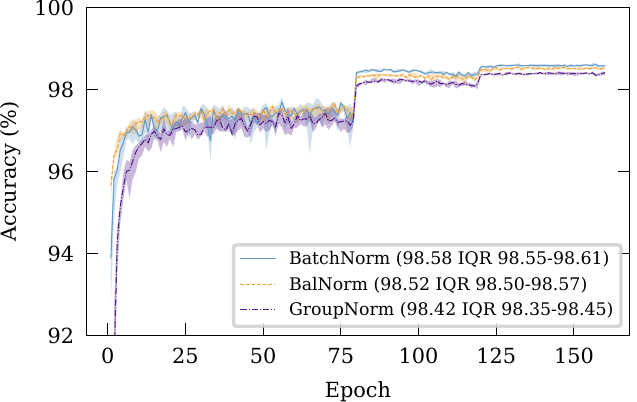}

\includegraphics{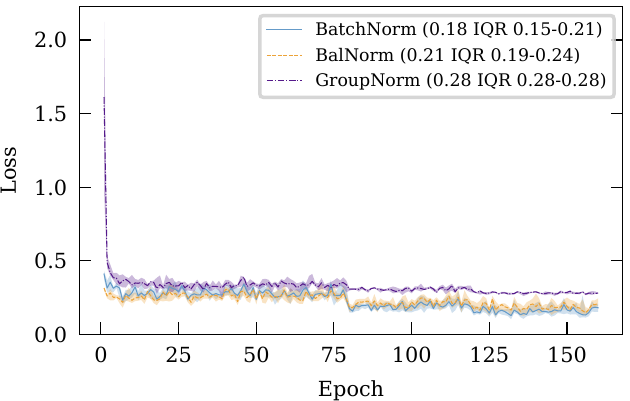}}
\par\end{centering}
\begin{centering}
\subfloat[\label{fig:SVHN-1}ILSVRC 2012 ImageNet (ResNet-50)]{\includegraphics{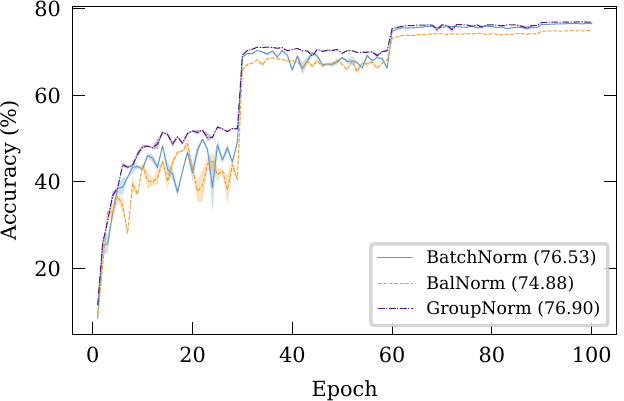}

\includegraphics{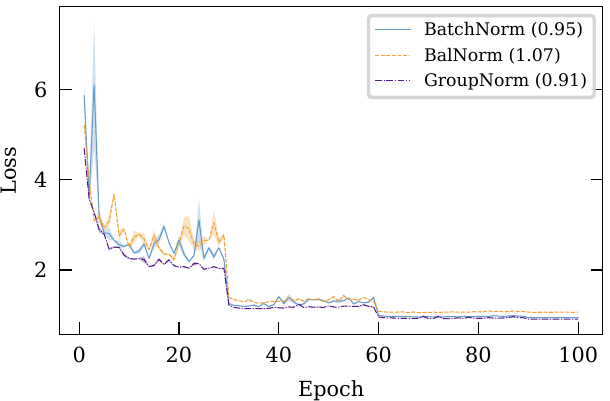}}
\par\end{centering}
\begin{centering}
\caption{Test set accuracy and loss. Median of 10 runs shown with interquartile
regions overlaid with the exception of the ImageNet plot which uses
3 runs.}
\par\end{centering}
\end{figure}

The SVHN+EXTRA dataset \citep{svhn} is much larger than CIFAR-10/100
(73,257 + 531,131 training instances), so we trained across 2 GPUs,
using $2\times$ larger mini-batches (size 256) so as to keep the
batch statistics noise (which are computed on a per-gpu basis) the
same. Other hyper-parameters were also kept the same, with the exception
that we trained for fewer epochs (with LR reductions moved to epochs
80 and 120). Figure \ref{fig:SVHN} shows that BalNorm achieves essentially
the same generalization performance as BatchNorm. On this problem
GroupNorm appears inferior, although this may be due to the default
group-size of 32 being suboptimal here.

\subsection{ILSVRC 2012 ImageNet}

We also ran some preliminary experiments on the ILSVRC 2012 ImageNet
classification task using the standard ResNet50 architecture \citep{resnet}.
Our results here show a generalization gap between BalNorm and BatchNorm/GroupNorm.
It may be possible to eliminate this gap using additional regularisation
as in the CIFAR-10 case, however we found manifold mixup to not yield
such an improvement.

\subsection{Reporting training variability}

We are careful to report results aggregated over enough runs involving
different RNG seeds so that run-to-run variability does not effect
our conclusions. This is absolutely necessary for the smaller test
problems above as the differences between runs can be comparable to
the difference between the compared normalization methods, and indeed
differences in reported results in the literature. We report median
and inter-quartile statistics (i.e. our plots include point-wise 25\%
and 75\% percentile ranges of the values seen), as these are more
representative of actual performance, and particularly the asymmetry
of test accuracy variability. The more commonly used two-standard-deviation
bars based on a normal assumption can show values both below and above
actually seen data (such as $>100\%$ accuracy upper bounds), and
are not supported by statistical theory for small samples such as
the ten used here.

\textbf{\bibliographystyle{plainnat}
\bibliography{norm}
 }
\end{document}